\documentclass[letterpaper, 10 pt, journal, twoside]{IEEEtran}
\usepackage{booktabs}
\usepackage{graphicx}
\usepackage{multirow}
\usepackage{bbding}
\usepackage{color}
\usepackage{url}
\usepackage{amsmath,amssymb}
\usepackage[colorlinks,linkcolor=red,anchorcolor=blue,citecolor=green]{hyperref}
\newcommand{\orcid}[1]{\href{https://orcid.org/#1}{\includegraphics[width=10pt]{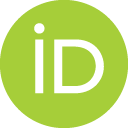}}}

\IEEEoverridecommandlockouts

\title{StreamMOS: Streaming Moving Object Segmentation with Multi-View Perception and Dual-Span Memory
}

\author{Zhiheng Li\orcid{0000-0002-1477-2066}, Yubo Cui\orcid{0000-0001-5302-0484}, Jiexi Zhong\orcid{0009-0000-3388-7865}, and Zheng Fang*\orcid{0000-0003-3887-3141}
\thanks{Manuscript received: July, 25, 2024; Revised November, 13, 2024; Accepted December, 5, 2024.} 
\thanks{This paper was recommended for publication by Editor Cesar Cadena Lerma upon evaluation of the Associate Editor and
Reviewers’ comments. This work was supported in part by the National Natural Science Foundation of China under Grants 62073066, in part by the Fundamental Research Funds for the Central Universities under Grant N2226001, and in part by 111 Project under Grant B16009. (\textit{Corresponding author: Zheng Fang})}
\thanks{The authors are all with Faculty of Robot Science and Engineering, Northeastern University, Shenyang 110819, China. Zhiheng Li and Zheng Fang are also with the National Frontiers Science Center for Industrial Intelligence and Systems Optimization, Northeastern University, Shenyang 110819, China and also with Key Laboratory of Data Analytics and Optimization for Smart Industry, Ministry of Education, Northeastern University, Shenyang 110819, China. (e-mail: fangzheng@mail.neu.edu.cn)}
\thanks{The code will be open at https://github.com/NEU-REAL/StreamMOS.git.}
\thanks{Digital Object Identifier (DOI): see top of this page.}
}

\begin{document}

\maketitle

%%%%%%%%%%%%%%%%%%%%%%%%%%%%%%%%%%%%%%%%%%%%%%%%%%%%%%%%%%%%%%%%%%%%%%%%%%%%%%%%
\begin{abstract}
Moving object segmentation based on LiDAR is a crucial and challenging task for autonomous driving and mobile robotics. Most approaches explore spatio-temporal information from LiDAR sequences to predict moving objects in the current frame. However, they often focus on transferring temporal cues in a single inference and regard every prediction as independent of others. This may lead to inconsistent segmentation results for the same object across different frames. To solve this issue, we propose a streaming network with a memory mechanism, called StreamMOS, to build the association of features and predictions among multiple inferences.
%%%%%%
Specifically, we utilize a short-term memory to convey historical features, which can be regarded as spatial priors of moving objects and are used to enhance current inference by temporal fusion. 
Meanwhile, we build a long-term memory to store previous predictions and exploit them to refine current forecasts at the voxel and instance levels through voting. Besides, we apply multi-view encoder with cascaded projection and asymmetric convolution to extract motion feature of objects in different representations. 
Extensive experiments validate that our algorithm gets competitive performance on SemanticKITTI and Sipailou Campus datasets. 
\end{abstract}

\begin{IEEEkeywords}
Semantic Scene Understanding; Deep Learning Methods; Computer Vision for Transportation
\end{IEEEkeywords}

%%%%%%%%%%%%%%%%%%%%%%%%%%%%%%%%%%%%%%%%%%%%%%%%%%%%%%%%%%%%%%%%%%%%%%%%%%%%%%%%
\vspace{-0.15in}
\section{Introduction}
\IEEEPARstart{O}{n} urban roads, there are often many dynamic objects with variable trajectories, such as vehicles and pedestrians, which create the collision risk for autonomous vehicles. Meanwhile, these moving objects will cause errors in simultaneous localization and mapping (SLAM)~\cite{loam}, as well as pose challenges for obstacle avoidance~\cite{obstacle} and path planning~\cite{deeppathplanning}. 
As a result, online moving object segmentation (MOS) based on LiDAR points has become a crucial task in multiple fields.
However, owing to the unordered and sparse nature of LiDAR points, MOS still faces some challenges, especially difficulty in perceiving moving objects with sparse points at a distance.

To tackle the above problem, the mainstream strategy is to exploit spatio-temporal information from LiDAR sequences. For instance, Chen et al.~\cite{LMNet} generate residual image in range view (RV), which reflects spatial positions of moving objects in each frame and can be utilized to perform temporal fusion to infer motion states of objects. Based on the RV projection in \cite{LMNet}, Sun et al.~\cite{MotionSeg3D} adopt motion-guided attention to better explore temporal motion cues from residual images.
Besides, some works~\cite{Limoseg,MotionBEV} attempt to map point clouds to bird’s eye view (BEV) and ensure consistent object size and movement. 
Recently, Wang et al.~\cite{insmos} process LiDAR sequences directly via 4D convolution to construct temporal associations while adding instance detection to promote segmentation integrity. 

\begin{figure}[t]
\centering
\includegraphics[width=\linewidth, height=7.6cm]{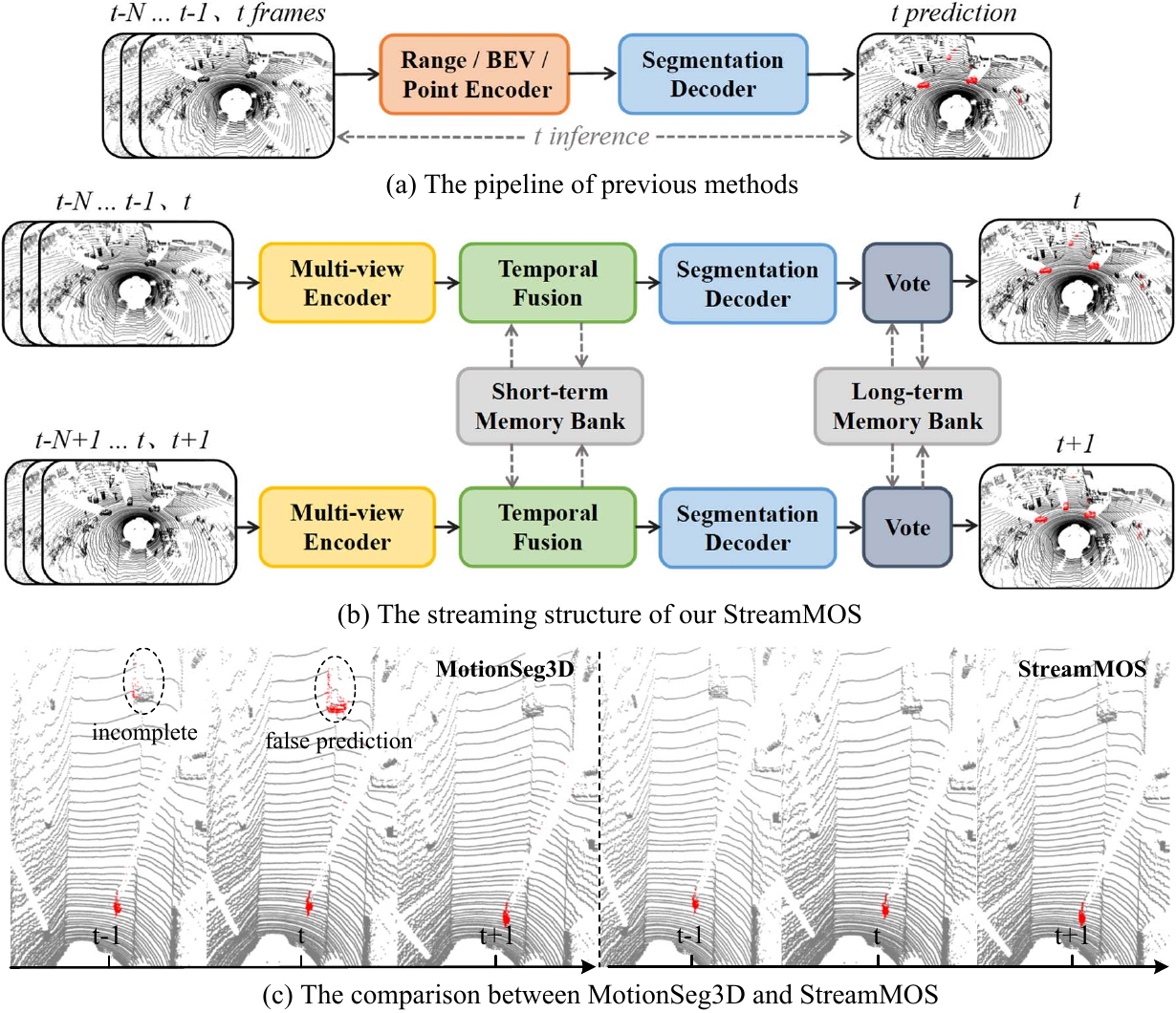}
\vspace{-0.26in}
\caption{\textbf{Pipeline comparison of moving object segmentation approaches.} We compare the structure of proposed StreamMOS with previous methods in (a) and (b). Meanwhile, the segmentation results obtained by our method achieve better spatial integrity and temporal continuity in (c).}
\label{fig:Compare Framework}
\vspace{-0.2in}
\end{figure} 

However, as displayed in Fig.~\ref{fig:Compare Framework}(a), these methods focus on temporal fusion in a single inference and make independent predictions for each frame, leading to inconsistent results for the same object at different moments (in Fig.~\ref{fig:Compare Framework}(c)).
Despite Mersch et al.~\cite{4DMOS} leveraging a binary Bayes filter to combine multiple predictions, it still ignores information transmission at feature level, which supplies rich spatial context to the next inference. 
%%%%%
Thus, we present a “streaming” structure as shown in Fig.~\ref{fig:Compare Framework}(b), which regards historical feature as a strong prior and exploits it to guide the current inference. Meanwhile, the past predictions are stored in long-term memory and utilized to suppress false predictions. 
In this way, we construct robust correlations in multiple inferences and fully explore temporal information to ensure consistent results in different frames.

To implement the idea of streaming, we propose a moving object segmenter, named \textit{StreamMOS}, which captures object motion across multiple views and adopts dual-span memory to transfer historical information. 
%%%%%%%%%
Specifically, different from previous works that map point clouds to one view, we argue that various viewpoints provide more holistic observations of dynamic objects.
%%%%%%%%%
Thus, we propose a multi-view encoder that uses a cascaded structure to iteratively get dense appearance from RV and perceive intuitive motion on BEV, resulting in more discriminative features of moving objects (in Fig.~\ref{fig:MVE}(b)). 
%%%%%%%%%
Meanwhile, during BEV encoding, we introduce asymmetric convolution with decoupled strategy to better capture vertical and horizontal motion.
%%%%%%%%%
Then, we use attention mechanism to implement temporal fusion that aligns features from different times and conveys spatial prior to current inference. 
%%%%%%%%%
Besides, due to the inherent uncertainty of neural networks, the output of segmentation decoder could be inconsistent across frames (Fig.~\ref{fig:Compare Framework}(c)). To solve this issue, we propose voting mechanism as post-processing to optimize predicted labels.
Its core idea is to statistically analyze long-term motion states at voxel and instance levels, and then select the most likely state to update raw point-wise forecasts.
In this way, the previous results can be used to refine current predictions, enhancing the temporal continuity and spatial completeness of segmentation together.

In summary, the contributions of our work are as follows: 
\begin{itemize}
\item We present a novel streaming framework called StreamMOS, which exploits short-term and long-term memory to construct associations among inferences and improve the integrity and continuity of predictions in MOS task.
\item We propose a multiple projection architecture to capture object motion and complete appearance from different views. We introduce a multi-level voting mechanism to refine segmentation results for every voxel and instance.
\item The experimental results confirm that our StreamMOS outperforms the previous state-of-the-art on the test sets of SemanticKITTI by 1.1\% IoU and Sipailou Campus by 1.7\% IoU, while achieving competitive running time.
\end{itemize}

%%%%%%%%%%%%%%%%%%%%%%%%%%%%%%%%%%%%%%%%%%%%%%%%%%%%%%%%%%%%%%%%%%%%%%%%%%%%%%%%
\section{Related Work}

\subsection{Geometric-based Algorithms}

The initial LiDAR-based MOS algorithms can be referred to as the \textit{geometric-based approaches}, which typically build the map in advance and remove any dynamic objects through estimating occupancy probability and determining visibility.
%%%%%%%%%%%%
For example, Schauer et al.~\cite{Peopleremover} proposed a ray casting-based approach that counted the hits and misses of scans to update the occupancy situation of the grid map. Afterwards, Pagad et al.~\cite{Robust} constructed an occupancy octree map and proposed a probability update mechanism to obtain clean point clouds by considering the occupancy history.  
%%%%%%%%%%%%
Despite getting promising results, \cite{Peopleremover, Robust} suffer extensive computational burden due to the ray casting and updating voxel one by one. To improve efficiency, several visibility-based~\cite{Long-term,Remove,Meta-rooms} algorithms have been developed. Pomerleau et al.~\cite{Long-term} identified moving objects by checking whether the points of the pre-built map are occluded by the points in the query frame. Meanwhile, to avoid mismarked ground points as dynamic reported in~\cite{Long-term}, Kim et al.~\cite{Remove} retained ground points from removed points using a multi-resolution reverting algorithm.
%%%%%%%%%%%%
Moreover, Lim et al. \cite{ERASOR} introduced a visibility-free approach that removed 
moving traces by computing pseudo occupancy ratio between the query scan and the submap in each grid. 
%%%%%%%%%%%%
Building on~\cite{ERASOR}, Zhang et al. \cite{erasor++} proposed a height coding descriptor, while Lim et al. \cite{erasor2} introduced instance segmentation to maximize the preservation of static points.
%%%%%%%%%%%%
Although the above methods clean maps well, they are performed offline due to requiring a prior map, making them unsuitable for real-time applications.
%%%%%%%%%%%%

\subsection{Learning-based Algorithms}
Recently, many studies have focused on utilizing \textit{learning-based approaches} to eliminate dynamic objects online, which take only consecutive frame point clouds as input rather than a pre-built map. Meanwhile, according to data representation, these algorithms can be categorized into projection-based and point-based methods. The former converts point clouds into bird’s eye view (BEV) or range view (RV) images, while the latter processes 3D raw points directly.

Specifically, for \textit{point-based algorithms}, Mersch et al.~\cite{4DMOS} adopted sparse 4D convolutions to process a series of LiDAR scans and predicted moving objects in each frame. They also employed a binary Bayes filter to fuse multiple predictions in a sliding window. 
%%%%%%%%%%%%
Subsequently, Kreutz et al.~\cite{Unsupervised4Dlidar} proposed an unsupervised approach to address MOS task in stationary LiDAR and viewed it as a multivariate time series clustering problem.
%%%%%%%%%%%%
Lately, Wang et al.~\cite{insmos} introduced InsMOS to unify detection and segmentation of moving objects into a network, so that the instance cues can be used to improve segmentation integrity. Although they achieved promising performance, the feature extraction of numerous points in~\cite{insmos} may lead to high computational costs. 

Compared to the mentioned approaches, \textit{projection-based algorithms}~\cite{LMNet,MotionSeg3D,Rvmos,Limoseg,MotionBEV} are generally more efficient owing to handling ordered and dense data.
For instance, Chen et al.~\cite{LMNet} mapped LiDAR scans into spherical coordinates and generated residual images to extract dynamic information in sequence. Sun et al.~\cite{MotionSeg3D} used a dual-branch network to encode spatio-temporal information and mitigated boundary blurring problem with a point refinement module.
%%%%%%%%%%%%
In addition, Kim et al.~\cite{Rvmos} achieved higher performance by using extra semantic features. 
In contrast to the RV projection, Mohapatra et al.~\cite{Limoseg} and Zhou et al.~\cite{MotionBEV} utilized BEV projection to obtain a more intuitive motion representation, but the serious loss of spatial information still limited performance.
To solve this issue, our StreamMOS captures object motion from multiple views in a series manner, allowing for complete observation of objects.
We also build memory banks to convey historical knowledge, resulting in consistent segmentation across a long sequence.

\begin{figure*}[t]
\centering
\includegraphics[width=16.8cm, height=8.3cm]{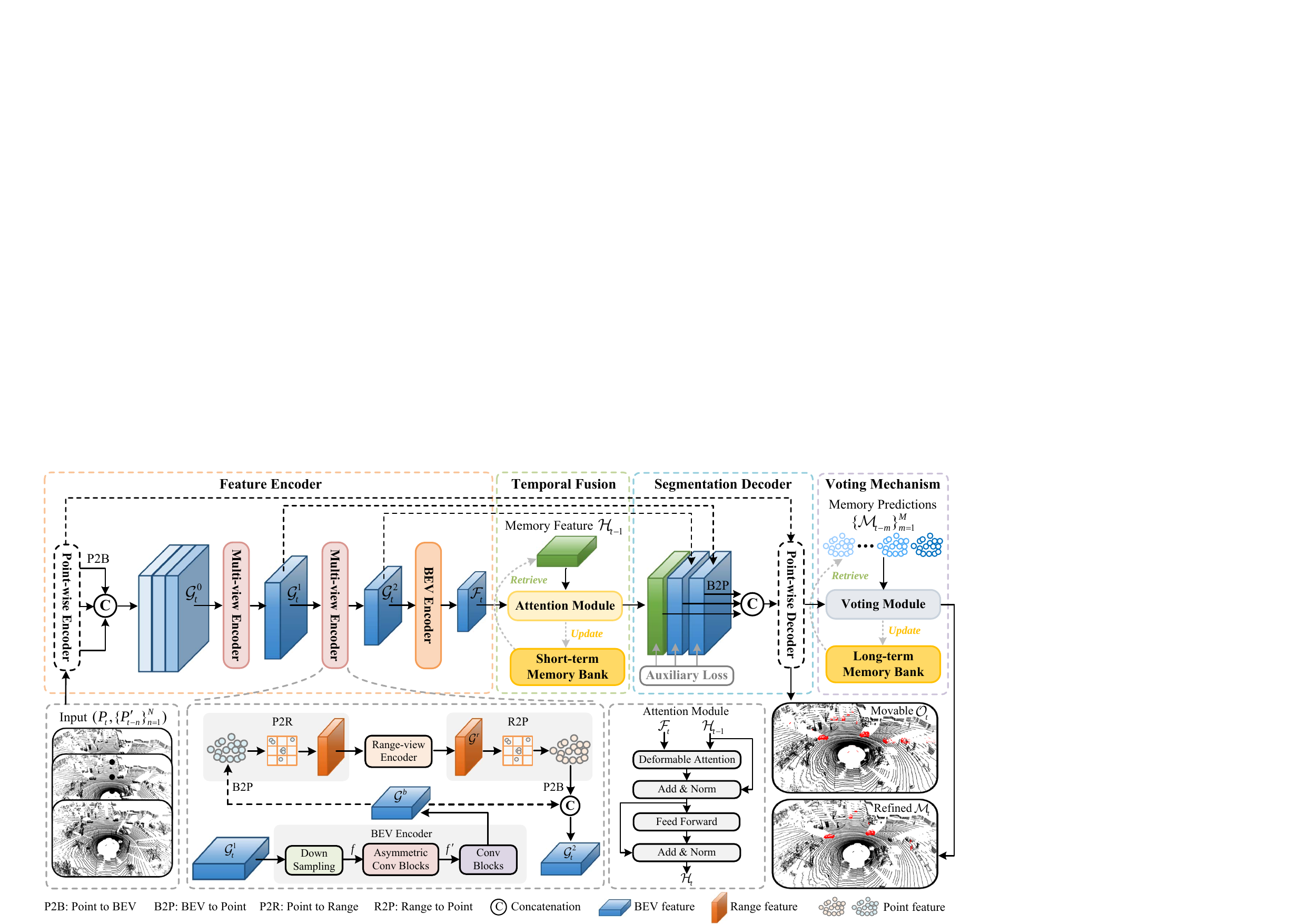}
\vspace{-0.18in}
\caption{\textbf{The overall architecture of StreamMOS.} (a) Feature encoder adopts a point-wise encoder to extract point features and project them into BEV. Then, the multi-view encoder with cascaded structure and asymmetric convolution is applied to encode motion features from different views. (b) Temporal fusion utilizes an attention module to propagate memory feature to the current inference. (c) Segmentation decoder with parameter-free upsampling exploits multi-scale features to predict class labels. 
(d) Voting mechanism leverages memory predictions to optimize the motion state of each 3D voxel and instance.
} 
\label{fig:StreamMOs}
\vspace{-0.18in}
\end{figure*} 

%%%%%%%%%%%%%%%%%%%%%%%%%%%%%%%%%%%%%%%%%%%%%%%%%%%%%%%%%%%%%%%%%%%%%%%%%%%%%%%%
\section{Methodology}
\subsection{Framework Overview}
LiDAR-based MOS aims to determine the motion state of each point in the current scan based on the multi-frame point clouds $\{\mathcal{P}_{t-n}\}_{n=0}^{N}$. To this end, existing methods first adopt the relative pose transformations $\{\mathcal{T}_{t-n \rightarrow t}\}_{n=1}^{N}$ provided by the LiDAR odometry to project historical scans $\{\mathcal{P}_{t-n}\}_{n=1}^{N}$ into ego car coordinate system of the current scan $\mathcal{P}_t$ and get $\{\mathcal{P}'_{t-n}\}_{n=1}^{N}$. Then, they usually feed $\mathcal{P}_t$ and $\{\mathcal{P}'_{t-n}\}_{n=1}^{N}$ into a network $\Psi$ to fuse spatio-temporal information and predict classification results $\mathcal{M}_{t} \in \mathbb{R}^{V \times 3}$ for all points in $\mathcal{P}_t$, where $V \times 3$ refers to the probability that $V$ points belong to three categories, including unknown, static and moving states. 

Different from previous approaches that focus on temporal fusion in a single inference, we extra consider the association among multiple inferences and apply historical feature $\mathcal{H}_{t-1}$ and predictions $\{\mathcal{M}_{t-m}\}_{m=1}^{M}$ to raise the quality of current inference. Thus, our method formulates MOS task as follows:
\begin{align}
\mathcal{M}_{t} = \Psi (\mathcal{P}_t, \{\mathcal{P}'_{t-n}\}_{n=1}^{N}, \mathcal{H}_{t-1}, \{\mathcal{M}_{t-m}\}_{m=1}^{M})
\end{align}
where $N, M$ are the number of historical LiDAR frames and forecasts. Meanwhile, the details of our network are shown in Fig.~\ref{fig:StreamMOs}. Specifically, given a series of scans, our StreamMOS first leverages the multi-view encoder to capture motion cues from the viewpoints of BEV and RV. Thereafter, we can get a motion feature $\mathcal{F}_{t}$ that reflects spatial information of moving objects in the current frame.
Then, we use a temporal fusion module to combine $\mathcal{F}_{t}$ with historical feature $\mathcal{H}_{t-1}$ retained in short-term memory. By doing this, some prior information can be transferred to the current inference and further utilized to decode movable objects $\mathcal{O}_{t}$ and coarse motion state $\mathcal{C}_{t}$ for all points. Finally, we employ a voting mechanism to update $\mathcal{C}_{t}$ using historical results $\{\mathcal{M}_{t-m}\}_{m=1}^{M}$ stored in long-term memory and instance information derived from $\mathcal{O}_{t}$, thereby yielding the refined prediction $\mathcal{M}_{t}$.

\subsection{Multi-projection Feature Encoder}
\subsubsection{Preliminaries} 
Unlike the existing methods that project point clouds into a single view, such as BEV~\cite{MotionBEV} or RV~\cite{Rvmos}, we believe that mapping points to these views simultaneously could capture more complete appearance and obvious motion cues of dynamic objects. Meanwhile, as shown at the bottom of Fig.~\ref{fig:StreamMOs}, the points could be considered as the intermediate carrier to transfer information between different perspectives.
For this purpose, we adopt \textit{Point-to-BEV} (P2B) and \textit{Point-to-Range} (P2R) to map point features into 2D plane, while using \textit{BEV-to-Point} (B2P) and \textit{Range-to-Point} (R2P) to collect the point features from multiple planes. 
%%%%%%%%%
To be specific, assuming that the $k^{th}$ 3D point in $\mathcal{P}_{t}$ is denoted as $p_k^{3D}=(x_k,y_k,z_k)$, the P2B projects it into a rectangular 2D grid and obtains its coordinate $(u_k^b, v_k^b)$ in BEV. For the P2R, the point $p_k^{3D}$ with 3D cartesian coordinate is converted into spherical coordinate $p_k^{sph} = (r_k, \theta_k, \phi_k)$ and assigned to the 2D grid in RV with coordinate $(u_k^r, v_k^r)$~\cite{CPGNet}, where $r_k$, $\theta_k$, $\phi_k$ represent distance, zenith and azimuth angle of point $p_k^{3D}$. 
The points falling into the same grid undergo max-pooling to aggregate features. For R2P and B2P, the grid features of RV and BEV are allocated to 3D points using bilinear interpolation within nearby grids.

\subsubsection{Network Structure} In the feature encoder, we first use a lightweight PointNet~\cite{PointNet} as point-wise encoder to process point clouds ($\mathcal{P}_{t}, \{\mathcal{P}'_{t-n}\}_{n=1}^{N}$) and obtain \( \mathcal{E}_n \in \mathbb{R}^{V \times C} \ (n \in \{t-N,...,t \}) \), where $C$ means the number of channels. Then, for the feature of each frame, we adopt P2B to project them into BEV and concatenate them along the channel dimension to get BEV feature $\mathcal{G}_t^0 \in \mathbb{R}^{W^b \times H^b \times (N+1)C}$, where $W^b, H^b$ are the predefined width and height of BEV. Afterwards, we feed $\mathcal{G}_t^0$ into multi-view encoder (MVE) to extract temporal information and capture object motion from different views.

\begin{figure}[t]
\centering
\vspace{0.05in}
\includegraphics[width=8.0cm, height=6.1cm]{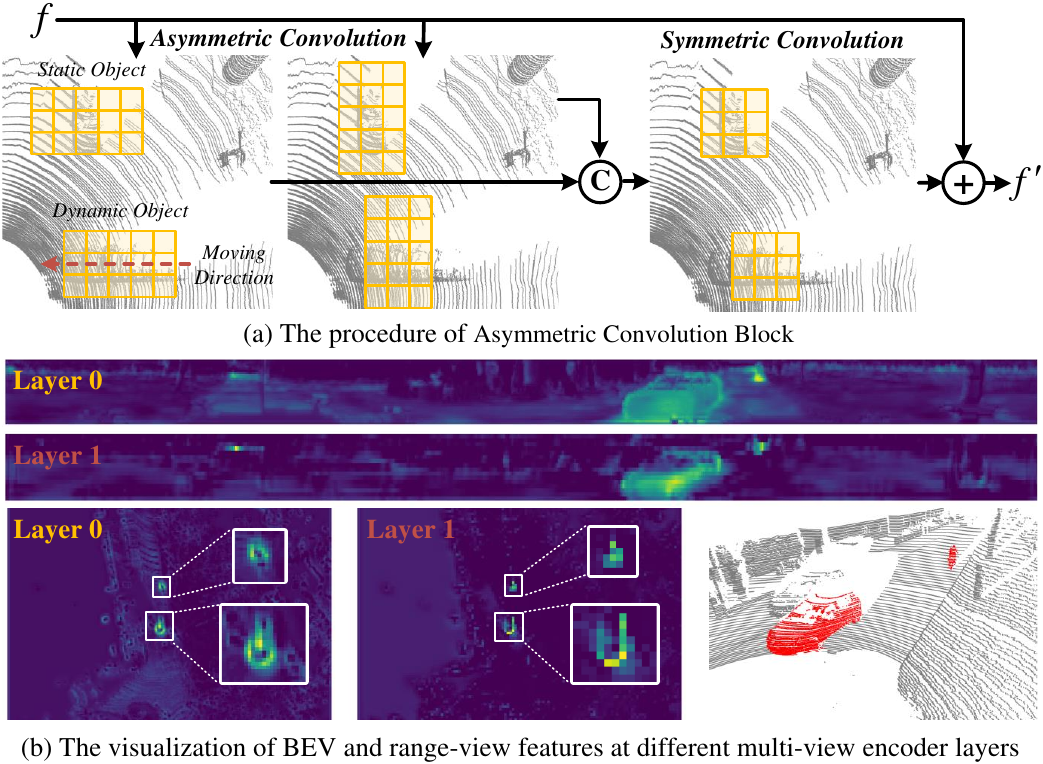}
\vspace{-0.12in}
\caption{\textbf{Illustration of asymmetric convolution and multi-view features.}}
\label{fig:MVE}
\vspace{-0.23in}
\end{figure}

In the lower part of Fig.~\ref{fig:StreamMOs}, after downsampling BEV feature $\mathcal{G}_t^l (l \in \{0, 1\})$, we introduce an asymmetric convolution block (ACB) to perceive the movement of objects. As shown in Fig.~\ref{fig:MVE}(a), compared to the typical symmetric convolutional kernel (e.g., 3$\times$3), the kernel size of ACB has one side longer (e.g., 3$\times$5 and 5$\times$3). Besides, it decouples feature extraction into the horizontal and vertical directions, defined as follows:
\begin{align}
f' = \text{Conv}_{3\times 3}(\text{Conv}_h(f) \odot \text{Conv}_v(f)) + f
\end{align}
where $f$ and $\odot$ are feature map and concatenation operation. $\text{Conv}_h$ and $\text{Conv}_v$ mean asymmetric convolutions, which can expand the receptive field and improve perception ability for moving objects since they usually have distinct motion in a specific direction. 
After that, as displayed in Fig.~\ref{fig:StreamMOs}, we apply B2P and P2R to project BEV feature into the range view and then use convolution layer as range-view encoder to generate another motion feature $\mathcal{G}^r$, which is remapped into BEV and combined with $\mathcal{G}^b$ along channel dimension. The multi-view feature interaction is then executed in the next encoder layer.

As a result, complete motion information can be extracted through cascaded projection and encoding within two MVEs and an additional BEV encoder, thereby obtaining a discriminative motion feature $\mathcal{F}_{t}$.
Specially, we illustrate multi-view features of different MVE layers in Fig.~\ref{fig:MVE}(b). It proves that MVE can extract consistent object information across various perspectives, while the deeper layer is capable of suppressing noise and preserving clearer motion features.

\subsection{Short-term Temporal Fusion}
The purpose of this part is to transfer the memory feature $\mathcal{H}_{t-1}$ from the last inference to the present, so that historical spatial states of objects can be retrieved to guide the network in inferring object motion at time $t$. 
To achieve this, we first build short-term memory bank as a bridge to store $\mathcal{H}_{t-1}$ and connect adjacent inferences.
Then, since $\mathcal{F}_{t}$ and $\mathcal{H}_{t-1}$ are not in the same coordinate system, we adopt an attention module with learnable offsets~\cite{DeformableDETR} to adaptively find the relationship between two features and combine them by attention weight.
Specifically, $\mathcal{H}_{t-1}$ is fed into two linear layers to produce $K$ attention weights $A_k$ and sampling offsets $\Delta g_k$. Later, based on the offsets $\Delta g_k$ and coordinates $g_k$ of reference points in $\mathcal{F}_{t}$, a bilinear interpolation is used to gather reference values $G_k$ from $\mathcal{F}_{t}$.
Finally, $G_k$ is weighted by $A_k$ to get enhanced feature ${\hat{\mathcal{H}}_{t}}$. The above process can be formulated as follows:
\begin{equation}
A_k = \text{Softmax}(\text{Linear}(\mathcal{H}_{t-1})), \ \Delta g_k = \text{Linear}(\mathcal{H}_{t-1})
\end{equation}
\begin{equation}
G_k = S(\mathcal{F}_{t}, g_k + \Delta g_k)
\end{equation}
\begin{equation}
{\hat{\mathcal{H}}_{t}} = \sum_{l=1}^{L} W_l (\sum_{k=1}^{K} A_{lk} \cdot G_{lk})     
\end{equation}
The $L$ and $K$ are the number of attention heads and reference points, respectively, while $S(·\cdot·)$ and $W_l$ represent the bilinear sampling and learnable weight of multi-head attention. Next, the ${\hat{\mathcal{H}}_{t}}$ is processed by normalization layer and feed-forward network (FFN) to generate a updated $\mathcal{H}_{t}$ at the current time:
\begin{equation}
\tilde{\mathcal{H}}_{t} = \text{LN}(\hat{\mathcal{H}}_{t} + \mathcal{H}_{t-1}), \ \mathcal{H}_{t} = \text{LN}(\text{FFN}(\tilde{\mathcal{H}}_{t}) + \tilde{\mathcal{H}}_{t})
\end{equation}
Here, LN is the layer normalization. Then, $\mathcal{H}_t$ is used for two purposes: it replaces $\mathcal{H}_{t-1}$ to update the short-term memory, and it is fed into the decoder to predict segmentation results.

\subsection{Reduced-parameter Segmentation Decoder}
To distinguish the static and dynamic points, the previous methods~\cite{MotionBEV,MotionSeg3D,insmos} usually leverage a UNet-like decoder to upsample multi-scale features progressively by convolutions. 
However, to reduce complexity and storage costs of network, we introduce a lightweight decoder, which first employs bilinear interpolation to convert the size of multi-scale features $\mathcal{G}_t^i \ (i \in {1,2})$ and $\mathcal{H}_t$ into a uniform height $H^b/2$ and width $W^b/2$. For each upsampled feature, we employ an auxiliary head to predict moving objects in BEV and exploit auxiliary loss as the constraint, which can guarantee that features from different scales are aligned and decoded well.
Next, we adopt B2P to convert upsampled features into point features one by one and combine them to get $ \mathcal{\hat{E}}_t \in \mathbb{R}^{V \times D}$.
Finally, in addition to decoding the coarse motion states $\mathcal{C}_{t} \in \mathbb{R}^{V \times 3}$ of LiDAR points $\mathcal{P}_{t}$, the point-wise decoder also outputs the probability $\mathcal{O}_{t} \in \mathbb{R}^{V \times 2}$  that points belong to movable objects (e.g., cars, bicycles) and static backgrounds (e.g., roads), represented as:
\begin{equation}
\mathcal{C}_{t} = \text{head}_1(\text{Conv}(\mathcal{\hat{E}}_t) \odot \mathcal{E}_t), \mathcal{O}_{t} = \text{head}_2(\text{Conv}(\mathcal{\hat{E}}_t) \odot \mathcal{E}_t)
\end{equation}where $\text{head}_1$ and $\text{head}_2$ consist of several convolution blocks, and $\mathcal{E}_t$ is the feature from point-wise encoder.
According to discrete classification labels $\mathcal{O}_{t}$, we can acquire the attributes of instance, like location and size, through clustering and use them to optimize $\mathcal{C}_{t}$ in the subsequent voting stage. 

\subsection{Long-term Voting Mechanism}
Most existing approaches~\cite{LMNet,MotionSeg3D,Rvmos} focus on improving the quality of a single inference through modifying network structure. Nevertheless, in light of the inexplicability and data dependency of neural networks, this strategy may be limited. For example, for a parked car shown in Fig.~\ref{fig:Compare Framework}(c), the model may predict it as stationary in one frame and moving in other frames. Meanwhile, due to lacking instance-level perception ability, the network may generate inconsistent results for the different parts of an object, particularly for cars (see Fig.~\ref{fig:vis}).

To solve these problems, we construct a long-term memory bank of length $M$ to store historical predictions and propose a voting module consisting of the voxel-based voting (VBV) and instance-based voting (IBV), which can function as post-processing to correct errors in the current predicted labels $\mathcal{C}_t$ using historical results $\{\mathcal{M}_{t-m}\}^M_{m=1}$ and movable labels $\mathcal{O}_{t}$.
Note that $\mathcal{M}_{t-m} \ (m=1,...,M)$ with coordinates of $\mathcal{P}_{t-m}$ is projected into the coordinate system of $\mathcal{P}_{t}$ to yield $\mathcal{M}'_{t-m}$ through the pose transformations $\mathcal{T}_{t-m \rightarrow t}$ in advance.

\subsubsection{Voxel-based voting} 
\begin{figure}[t!]
\centering
\includegraphics[width=\linewidth, height=6.2cm]{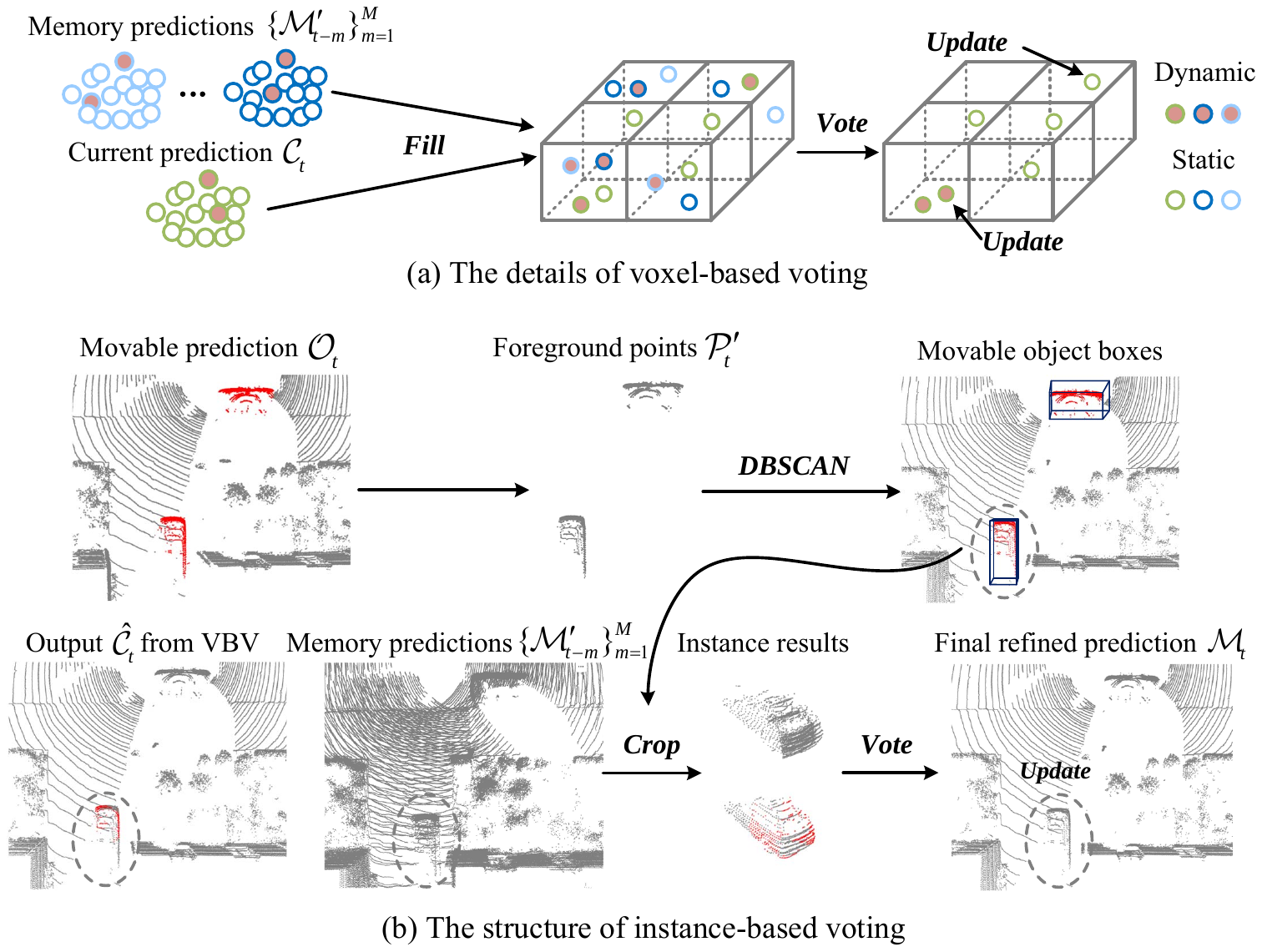}
\vspace{-0.3in}
\caption{\textbf{The details of our voting mechanism.} It uses voxel-based voting (VBV) and instance-based voting (IBV) to refine coarse predictions. }
\label{fig:vote mechanism}
\vspace{-0.20in}
\end{figure} 

Motivated by TFNet~\cite{TFNet}, we first obtain historical predictions $\{\mathcal{M}'_{t-m}\}^M_{m=1}$ and current forecast $\mathcal{C}_t$ in the same coordinate system. 
Then, we divide points $\mathcal{P}_{t}$ into voxels with fixed size and fill $(\mathcal{C}_t,\{\mathcal{M}'_{t-m}\}^M_{m=1})$ into each voxel. Next, as shown in Fig.~\ref{fig:vote mechanism}(a), the most frequently predicted label acts as motion state for all points in the same voxel and incorrect labels will be updated. We summarize the above procedure of VBV as: $\Omega (\mathcal{C}_{t}, \{\mathcal{M}'_{t-m}\}^M_{m=1}) \mapsto \hat{\mathcal{C}_{t}}$.
\subsubsection{Instance-based voting} 
Although VBV can ensure the consistency of motion states within a local area and achieve performance improvement in Tab.~\ref{tab: ablation model components}, it is difficult to achieve instance-level unity, as shown by the output $\hat{\mathcal{C}_{t}}$ from VBV in Fig.~\ref{fig:vote mechanism}(b). To solve this, we propose an instance-based voting based on clustering. When given the predicted probability $\mathcal{O}_{t}$ from decoder, we can pick out the foreground points ${\mathcal{P}'_{t}}$ from ${\mathcal{P}_{t}}$ and adopt DBSCAN~\cite{DBSCAN} to split ${\mathcal{P}'_{t}}$ into $S$ clusters.
Then, according to the coordinates of points in each cluster, we can compute $S$ minimum 3D bounding boxes to cover all objects. Thus, we can further crop out instance-level predictions from $\hat{\mathcal{C}_{t}}$ and memory predictions $\{\mathcal{M}'_{t-m}\}^M_{m=1}$. Finally, similar to voxel-based voting, we adopt the class label with the highest quantity as the motion state for all points in the instance and get the final prediction as: 
$\Phi(\hat{\mathcal{C}_{t}}, \{\mathcal{M}'_{t-m}\}^M_{m=1}, \mathcal{O}_{t}) \mapsto {\mathcal{M}_{t}}$.

Finally, when a new refined prediction $\mathcal{M}_t$ is output from voting mechanism, we append it to long-term memory while discarding the oldest result $\mathcal{M}_{t-M}$.
As a result, compared to relying on the network's adaptive learning, voting mechanism can explicitly suppress incorrect predictions and improve the consistency of segmentation based on a statistical analysis of historical predictions at the voxel and instance levels.

\subsection{Loss Functions}
\label{sec:Loss Functions}
To ensure the network can be fully optimized, we separate the training process into two steps. In the first stage, we only train our network without predicting movable objects in the decoder. Meanwhile, following the previous works~\cite{MotionBEV,MotionSeg3D}, we introduce the weighted cross-entropy ($L_{wce}$) and Lovász-Softmax ($L_{ls}$)~\cite{lovasz} losses to supervise network:
\begin{equation}
L = \lambda_1 L_{wce} + \lambda_2 L_{ls}, \ L_{s1} = L(y, \hat{y}) + \lambda_3 \sum_{i=1}^{3} L(y_i^b, \hat{y}_i^b)
\end{equation}
where $\lambda_{1}$, $\lambda_{2}$, $\lambda_{3}$ mean the weights for losses while $y$ and $\hat{y}$ are the ground truth and predicted results of points. $L(y_i^b, \hat{y}_i^b)$ denotes the auxiliary losses for BEV predictions. 
Moreover, in the second stage, we freeze pre-trained parameters that are optimized in the 1$^{st}$ stage and only train the rest of network to predict movable objects using the following loss function:
\begin{equation}
\ L_{s2} = \lambda_1 L_{wce}(x, \hat{x}) + \lambda_2 L_{ls}(x, \hat{x})
\end{equation}
where $x$ and $\hat{x}$ represent ground-truth labels and predictions for movable objects.

\begin{figure*}[t!]
\centering
\vspace{0.1in}
\includegraphics[width=\linewidth, height=7.0cm]{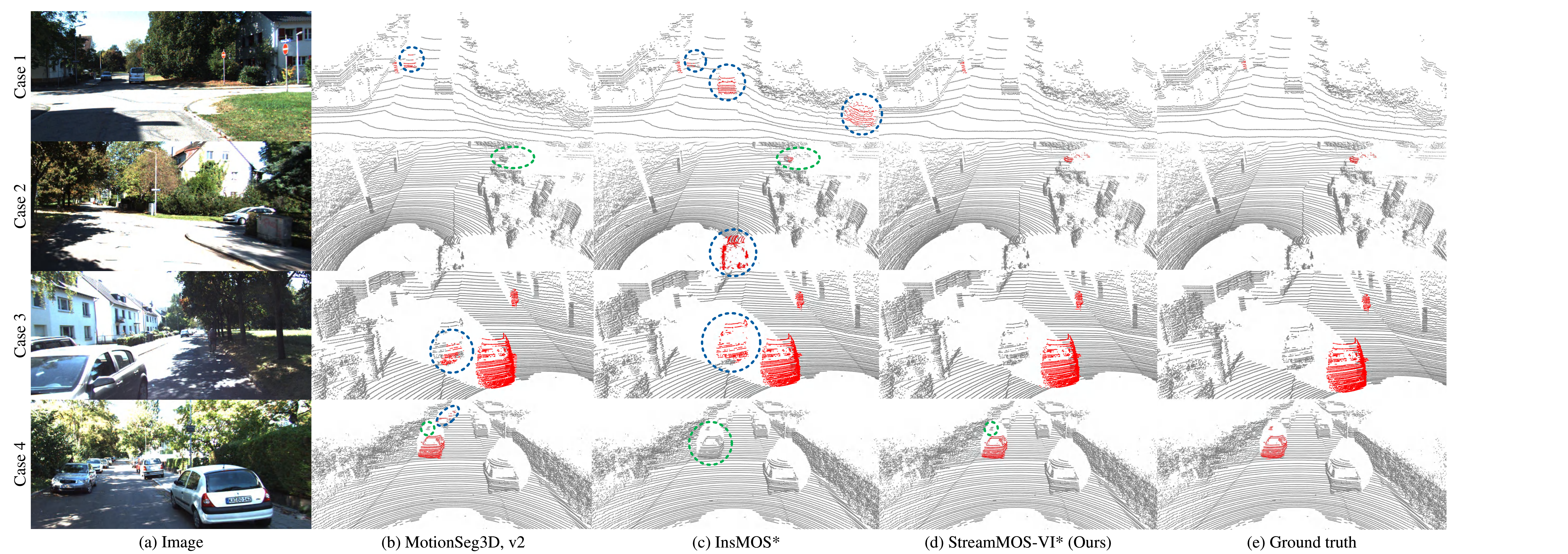}
\vspace{-0.3in}
\caption{\textbf{The visualization of MOS results on the SemanticKITTI validation set.} Incorrect predictions are highlighted, with false negatives marked by green circles and false positives by blue circles. Best viewed in color and zoom.} 
\label{fig:vis}
\vspace{-0.1in}
\end{figure*} 

\section{Experiments}
\subsection{Experimental Settings}
\noindent\textbf{Datasets.}
On SemanticKITTI-MOS~\cite{LMNet} dataset and Sipailou-Campus~\cite{MotionBEV} dataset, we compare segmentation performance with previous methods and conduct extensive ablation studies. The SemanticKITTI-MOS dataset is collected by a Velodyne HDL-64E LiDAR and contains a total of 22 sequences with labeled point clouds that are remapped from 28 semantic classes into 3 types of motion states. 
Following the previous algorithms~\cite{MotionBEV,MotionSeg3D,insmos}, we divide the sequences 00-07, 09-10 for training, sequence 08 for validation and sequences 11-21 for testing. For the Sipailou Campus dataset that is developed on solid-state LiDAR, we follow the implementation of~\cite{MotionBEV} to split dataset into 5 training sequences, 1 validation sequence and 2 test sequences from 26,279 frames.

\noindent\textbf{Evaluation Metric.}
Consistent with present approaches~\cite{insmos,MotionSeg3D}, we employ the Jaccard Index or Intersection-over-Union (IoU) metric~\cite{IOU} over dynamic objects to measure the MOS performance, which can be denoted as:
\begin{equation}
\text{IoU} = \frac{\text{TP}}{\text{TP}+\text{FP}+\text{FN}} 
\end{equation}
where TP, FP, and FN mean the number of true positive, false positive, and false negative predictions for dynamic category.
\subsection{Implementation Details}
In data processing, we leverage widely used data augmentation, such as random rotation, flipping and slight translation to enrich the training data, which plays an important role in improving model generalization. Meanwhile, as mentioned in Sec. \ref{sec:Loss Functions}, we optimize the network using two-stage training strategy. For the $1^{st}$ stage, we train the model for 48 epochs on NVIDIA RTX 4090 GPUs using an SGD optimizer with an initial learning rate of 0.02, which is decayed by 0.1 every 10 epochs. For the $2^{nd}$ stage, we solely optimize the network for 10 epochs with a learning rate of 0.02. 
Furthermore, each LiDAR scan is limited to [-50$m$, 50$m$] for the X and Y axes and [-4$m$, 2$m$] for the Z axis. The number of points in each scan is randomly downsampled or padded to $V$ = 1.3 $\times$ 10$^5$. 
The default values for the number of attention heads $L$ and reference points $K$ are both set to 4.
\subsection{Quantitative Results} 
\begin{table}[t]
\renewcommand\tabcolsep{8.5pt}
\vspace{-0.05in}
\caption{Performance comparison on SemanticKITTI validation and test sets. * denotes methods that exploit semantic labels. \dag \ indicates methods trained on both SemanticKITTI and KITTI-road datasets.}
\vspace{-0.05in}
~\label{tab:SemanticKITTI}
\centering
\scalebox{0.9}{
\begin{tabular}{c|c|c|c}
\toprule[.05cm]
Methods & Source & IoU (Validation) & IoU (Test) \\ \hline \hline
\rule{0pt}{8pt}
KPConv & \textit{ICCV 19} & - & 60.9 \\
SpSequenceNet & \textit{CVPR 20} & - & 43.2 \\
LiMoSeg & \textit{arXiv 21}  & 52.6 & - \\
LMNet & \textit{RA-L 21} & 66.4 & 58.3 \\
Cylinder3D & \textit{CVPR 21} & 66.3 & 61.2 \\
AutoMOS  & \textit{RA-L 22} & - & 54.3 \\
MotionSeg3D, v1 & \textit{IROS 22}  & 68.1 & 62.5 \\
MotionSeg3D, v2 & \textit{IROS 22}  & 71.4 & 64.9 \\
4DMOS  & \textit{RA-L 22}  & \underline{77.2} & 65.2 \\ 
MotionBEV, w/o delay & \textit{RA-L 23}  & 68.1 & 63.9 \\
MotionBEV, w/ delay & \textit{RA-L 23}  & 76.5 & \underline{69.7} \\
\textbf{StreamMOS-V} & -  & \textbf{78.3} & \textbf{73.1} \\ \hline
\rule{0pt}{8pt}
LMNet* & \textit{RA-L 21} & 67.1 & 62.5 \\
RVMOS* & \textit{RA-L 22}  & 71.2 & 74.7 \\ 
% LiDAR-BEVMTN* & \textit{arXiv 23}  & {73.1} & - \\
InsMOS* & \textit{IROS 23}  & 73.2 & 70.6 \\ 
InsMOS*\dag & \textit{IROS 23}  & - & 75.6 \\ 
MF-MOS* & \textit{ICRA 24}  & \underline{76.1} & \underline{76.7} \\
\textbf{StreamMOS-VI*} & -  & \textbf{81.6} & \textbf{77.8} \\
\toprule[.05cm]
\end{tabular}}
\vspace{-0.2in}
\end{table}
 
\noindent\textbf{Comparison with Previous Methods.} We first evaluate our StreamMOS on SemanticKITTI-MOS benchmark. To ensure fairness, our method is presented in two versions in Tab.~\ref{tab:SemanticKITTI} to make settings as consistent as possible with previous works. 
Specifically, (a) \textit{StreamMOS-V} indicates the network that is trained in the 1$^{st}$ stage and uses voxel-based voting as post-processing. (b) \textit{StreamMOS-VI*} means performing extra 2$^{nd}$ stage training and using instance-based voting that relies on movable object predictions. 
Then, our methods are compared with existing algorithms, which can be classified as whether semantic annotations are utilized. Specially, the two versions of MotionSeg3D~\cite{MotionSeg3D} refer to using kNN or point refinement as post-processing. Moreover, “w/ delay” signifies exploiting point cloud frames within the time window of $[t, t+N]$ to estimate dynamic objects in the $t$ frame. 
Note that following MotionBEV~\cite{MotionBEV}, our results shown in Tab.~\ref{tab:SemanticKITTI} are derived from training original SemanticKITTI without any additional data.

As illustrated in Tab.~\ref{tab:SemanticKITTI}, our streaming method outperforms previous works in most cases. Specifically, our StreamMOS-V exceeds 4DMOS~\cite{4DMOS} by 1.1\% and 7.9\% in validation and test. We think that compared with using binary Bayes filter to merge historical results in 4DMOS, our method additionally considers historical feature from the last inference, which can serve as strong spatial priors to improve prediction quality. At the same time, our StreamMOS-VI* surpasses InsMOS*~\cite{insmos} and MF-MOS*~\cite{MF-MOS} on the validation set significantly ($\uparrow$8.4\% and $\uparrow$5.5\%) by instance-based voting.
Finally, due to the lack of semantic annotation in Sipailou-Campus dataset, we solely list \cite{LMNet,MotionBEV,MotionSeg3D,4DMOS} in Tab.~\ref{tab:Sipailou Campus} and confirm the effectiveness of StreamMOS-V, even when using a solid-state LiDAR with a narrow field of view and non-repetitive scanning patterns.

\begin{table}[t]
\renewcommand\tabcolsep{10.0pt}
\vspace{-0.05in}
\caption{Performance comparison on Sipailou Campus dataset.}
\vspace{-0.05in}
~\label{tab:Sipailou Campus}
\centering
\scalebox{0.9}{
\begin{tabular}{c|c|c|c}
\toprule[.05cm]
Methods & Source & IoU (Validation) & IoU (Test) \\ \hline \hline
\rule{0pt}{8pt}
\rule{0pt}{8pt}
LMNet  & \textit{RA-L 21}  & 54.3 & 56.2 \\ 
MotionSeg3D, v2 & \textit{IROS 22}   & 65.6 & 66.8 \\
4DMOS & \textit{RA-L 22}   & 87.3 & 88.9 \\ 
MotionBEV & \textit{RA-L 23}  & \underline{89.2} & \underline{90.8} \\ 
\textbf{StreamMOS-V} & -  & \textbf{90.9} & \textbf{92.5} \\
\toprule[.05cm]
\end{tabular}}
\vspace{-0.15in}
\end{table}

\begin{table}[t!]
\renewcommand\tabcolsep{9pt}
\vspace{-0.05in}
\caption{Comparison of running times (ms) with previous methods.}
\vspace{-0.1in}
~\label{tab:Running Time}
\centering
\scalebox{0.9}{
\begin{tabular}{c|c|c|c}
\toprule[.05cm]
4DMOS & MF-MOS* & MotionSeg3D, v1 & MotionSeg3D, v2 \\ \hline 
\rule{0pt}{8pt}
86 & 96 & 42 & 117 \\ \hline
\rule{0pt}{8pt}
InsMOS* & RVMOS* & \textbf{StreamMOS-V} & \textbf{StreamMOS-VI*}\\ \hline
\rule{0pt}{8pt}
120 & 29 & 62 & 96 \\
\toprule[.05cm]
\end{tabular}}
\vspace{-0.25in}
\end{table}

\noindent\textbf{Inference Speed.} 
Although our method uses attention mechanism to construct feature association between inferences and merge multiple historical predictions in voting mechanism, it still keeps competitive running time compared with previous approaches in Tab.~\ref{tab:Running Time}. We believe this is the contribution of projection-based backbone, lightweight deformable attention and parameter-free upsampling in decoder, which enables our method strike a balance between speed and performance.

\begin{table}[t]
\renewcommand\tabcolsep{11.5pt}
\vspace{0.05in}
\caption{The effect of different modules in SemanticKITTI validation.}
\vspace{-0.06in}
~\label{tab:Component Ablation}
\centering
\scalebox{0.85}{
\begin{tabular}{c|cccc|cc}
\toprule[.05cm]
{} & {TF} & {MVE} & {VBV}  & {IBV} & IoU [\%] & $\Delta$ \\ \hline \hline
A1 & & & & & 67.1 & -  \\
A2 & \Checkmark & & & & 73.2 & \textcolor[rgb]{0.0,0.5,0.0}{+6.1} \\
A3 & \Checkmark & \Checkmark & & & 77.1 & \textcolor[rgb]{0.0,0.5,0.0}{+10.0} \\ 
A4 & \Checkmark & \Checkmark  & \Checkmark & & 78.3 & \textcolor[rgb]{0.0,0.5,0.0}{+11.2} \\ 
A5 & \Checkmark & \Checkmark & & \Checkmark & 81.3 & \textcolor[rgb]{0.0,0.5,0.0}{+14.2} \\
A6 & \Checkmark & \Checkmark  & \Checkmark & \Checkmark & \textbf{81.6} & \textcolor[rgb]{0.0,0.5,0.0}{+14.5} \\
\toprule[.05cm]
\end{tabular}}

\label{tab: ablation model components}
\vspace{-0.12in}
\end{table}

\begin{table}[t!]
\renewcommand\tabcolsep{11.0pt}
\caption{Ablation experiment on multi-view encoder of StreamMOS-V.}
\vspace{-0.09in}
~\label{tab:multi-view encoder}
\centering
\scalebox{0.85}{
\begin{tabular}{c|ccccc|c}
\toprule[.05cm]
{} & RV & BEV & ACB & Parallel & Series & IoU {[}\%{]} \\  \hline \hline
B1 & \Checkmark & & & & & 70.3 \\
B2 & & \Checkmark & & & & 74.2   \\
B3 & \Checkmark & \Checkmark & & & \Checkmark & 77.5  \\
B4 & \Checkmark & \Checkmark & & \Checkmark & & 74.8  \\
B5 & \Checkmark & \Checkmark & \Checkmark & & \Checkmark & \textbf{78.3} \\
\toprule[.05cm]
\end{tabular}}
\vspace{-0.12in}
\end{table}

\begin{table}[t!]
\renewcommand\tabcolsep{20.0pt}
\caption{Ablation experiment on temporal fusion of StreamMOS-V.}
\vspace{-0.09in}
~\label{tab:Detailed Ablation}
\centering
\scalebox{0.85}{
\begin{tabular}{c|c|c|c}
\toprule[.05cm]
{} & Strategy & IoU [\%]  & $\Delta$  \\ \hline \hline
C1 & w/o Temporal Fusion & 72.1 & \textcolor[rgb]{0.8,0.0,0.0}{-6.2} \\
C2 & Cross-attention & 73.0 & \textcolor[rgb]{0.8,0.0,0.0}{-5.3} \\
C3 & Concatenation & 74.8 & \textcolor[rgb]{0.8,0.0,0.0}{-3.5} \\
C4 & Addition & 75.6 & \textcolor[rgb]{0.8,0.0,0.0}{-2.7} \\
C5 & Deform-attention & \textbf{78.3} & - \\
\toprule[.05cm]
\end{tabular}}
\vspace{-0.25in}
\end{table}

\subsection{Qualitative Analysis} 
\begin{figure}[t!]
\centering
\vspace{0.05in}
\includegraphics[width=\linewidth, height=4.2cm]{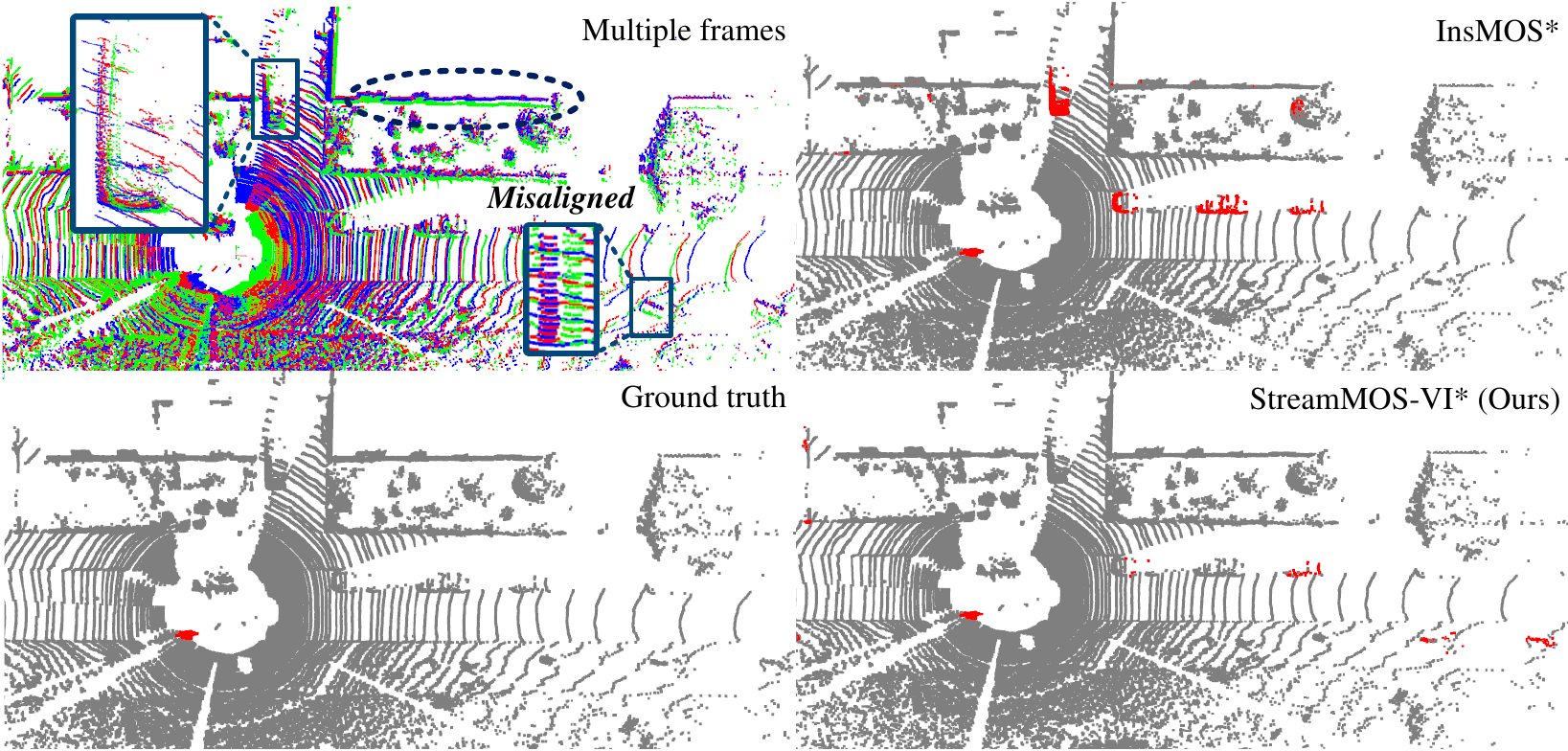}
\vspace{-0.27in}
\caption{\textbf{A failure case caused by inaccurate ego pose in SemanticKITTI.}}
\label{fig:failure}
\vspace{-0.13in}
\end{figure}

\noindent\textbf{Advantageous Cases.} 
In Fig.~\ref{fig:vis}, we exhibit the segmentation results in various scenarios to compare the previous methods intuitively. Although MotionSeg3D adopts a point refinement module to alleviate boundary-blurring problem, it still makes mistakes when dealing with distant objects, as shown in the 4$^{th}$ row. 
%%%%%%%%%%%%%%
Besides, MotionSeg3D tends to produce incomplete segmentation in the 3$^{rd}$ row due to lacking the instance-level sensing.
%%%%%%%%%%%%%%
Despite adding instance detection like InsMOS* can improve segmentation integrity, it aggravates negative impact when the prediction is incorrect, as illustrated in the 1$^{st}$, 2$^{nd}$ and 3$^{rd}$ rows.
%%%%%%%%%%%%%%
Unlike these algorithms, our StreamMOS-VI* combines multi-view observations to improve the perception of objects at different distances. Then, we build relationships among several inferences by integrating memory feature and predictions to enhance the segmentation integrity and reduce incorrect results. Thus, we get superior performance in Fig.~\ref{fig:vis}.
\noindent\textbf{Failure Cases.} Fig.~\ref{fig:failure} displays that the inaccurate ego poses $\{\mathcal{T}_{t-n \rightarrow t}\}_{n=1}^{N}$ misalign multi-frame point clouds, causing the network to incorrectly infer that the object has moved. While our method mitigates this issue compared to InsMOS*, errors persist as our voting mechanism still relies on precise poses. Thus, we think that developing a MOS model that eliminates the need for pose or implicitly learns ego-motion could be a promising research direction in the future.

\subsection{Ablation Study}
This part conducts ablation studies on the SemanticKITTI validation set to prove the effectiveness of our method.

\noindent\textbf{Model Components.} 
As shown in Tab.~\ref{tab:Component Ablation}, our StreamMOS mainly includes some crucial modules: temporal fusion (TF), multi-view encoder (MVE), voxel-based voting (VBV), and instance-based voting (IBV). To understand their importance in overall performance, we first remove all the above modules from our StreamMOS and regard the rest as a baseline in A1. 
After building feature correlations between inferences by TF, the IoU increases by 6.1\% in A2. Moreover, benefiting from capturing multi-view motion cues from BEV and RV, MVE brings further improvement. 
Then, due to introducing object-level perception, instance-based voting in A5 shows a greater performance than voxel-based in A4, which only focuses on limited areas in the 3D cube.
Finally, we can achieve optimal performance by combining them into a refinement procedure from voxel to instance, proving that effectively utilizing long-term predictions is the key element to the LiDAR MOS task. 

\noindent\textbf{Multi-view Encoder.} We compare several multi-view encoding strategies in Tab.~\ref{tab:multi-view encoder}. From the B1 and B2, we can observe that when encoding object motion only on a single view, the BEV representation achieves better results compared to RV due to global perspective and motion consistency.
Then, we divide encoder into BEV and RV branches and extract multi-view features in series (B3), leading IoU to further increase and exceed parallel mode (B4) by 2.7\%. We think that series manner may be more suitable for deriving consistent moving features from different views owing to progressive encoding.
Furthermore, using asymmetric convolution block (ACB) can result in 0.8\% improvement in B5, proving the advantage of decoupling horizontal and vertical encoding.

\begin{figure}[t!]
\centering
\vspace{0.05in}
\includegraphics[width=\linewidth, height=2.9cm]{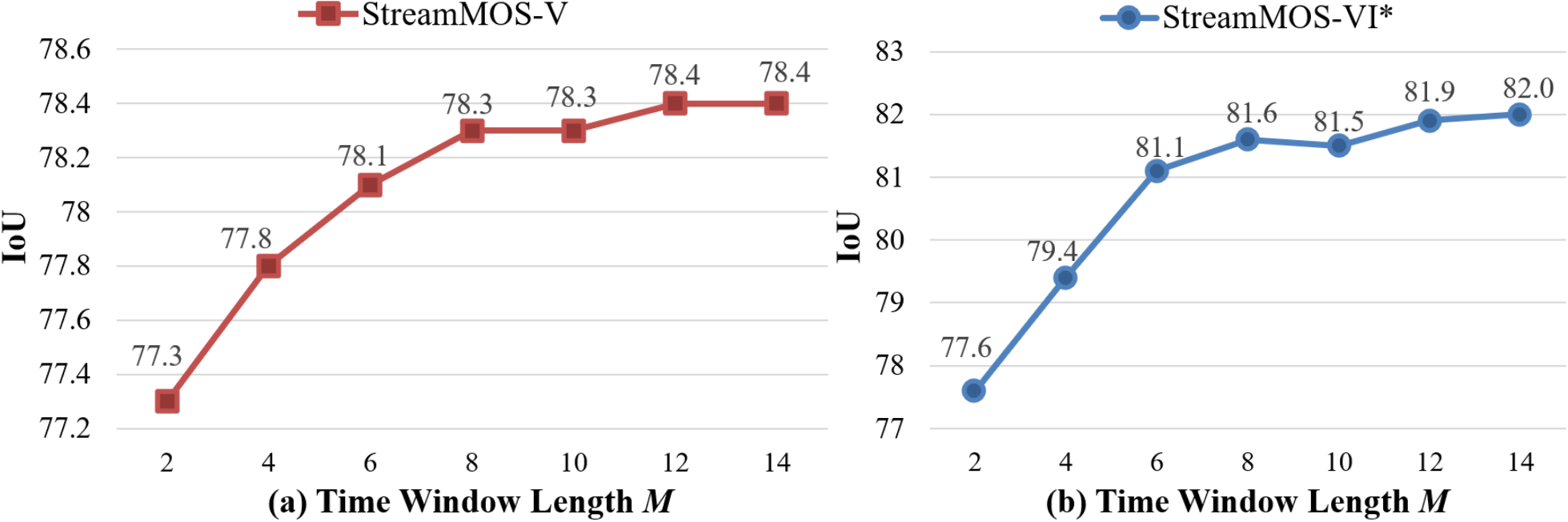}
\vspace{-0.27in}
\caption{\textbf{Ablation study on the time window length of voting mechanism.}}
\label{fig:time window length}
\vspace{-0.1in}
\end{figure} 

\begin{figure}[t!]
\centering
\includegraphics[width=\linewidth, height=2.6cm]{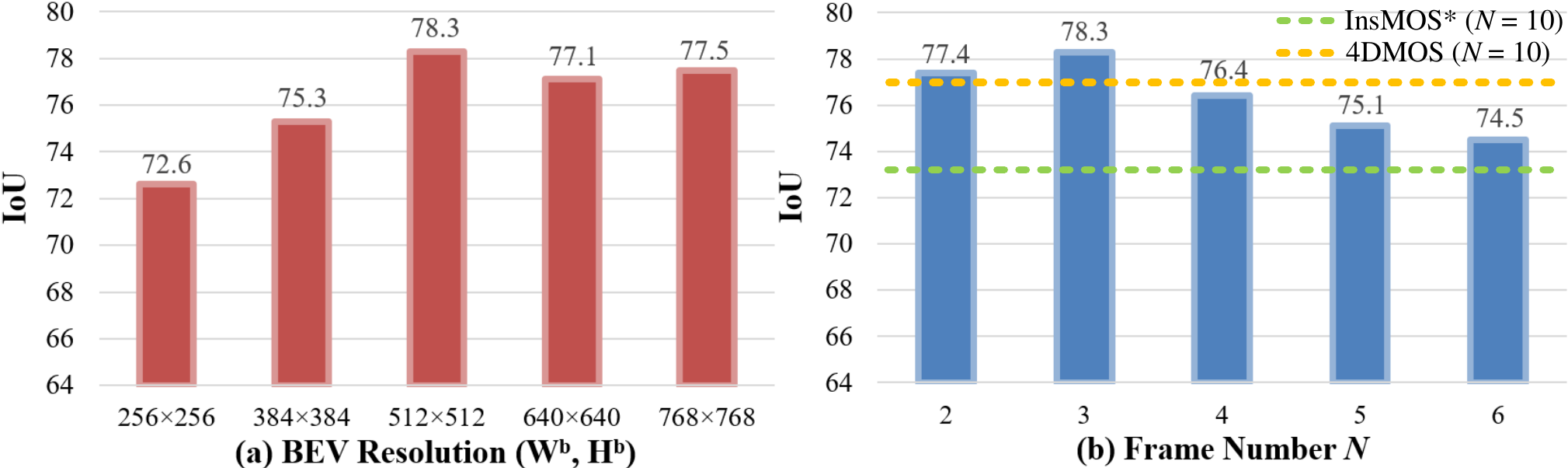}
\vspace{-0.27in}
\caption{\textbf{The effect of frame number and BEV size on the SreamMOS-V.}}
\label{fig:frame number and BEV size}
\vspace{-0.2in}
\end{figure}

\noindent\textbf{Temporal Fusion.} The strategy of propagating the historical feature into current inference will affect segmentation quality as demonstrated in Tab.~\ref{tab:Detailed Ablation}. 
First, we can observe that lacking temporal fusion to provide prior information leads to unideal results ($\downarrow$6.2\%). 
Then, compared with adopting concatenation and addition directly to merge features in different coordinate systems, deformable attention could align features adaptively by learnable offsets and gain the advantage of 3.5\% and 2.7\% IoU.
Moreover, it is worth noting that cross-attention gets the worst result since redundant global attention may cause a bad effect. In contrast, deformable attention concentrates on local feature to avoid model overfitting and save computation load.

\noindent\textbf{Time Window Length.} The time window length determines how long ago predictions can be used by voting mechanism. Thus, we conduct experiments on the time window length to choose the optimal setting for our algorithm. As displayed in Fig.~\ref{fig:time window length}, the performance will increase rapidly until the length $M$ of time window reaches 8.
Despite continuing to raise the length could result in a slight improvement, it requires more time consumption. Thus, we opt for $M$ = 8 as our default.

\noindent\textbf{Other Hyper-parameter Settings.} In Fig.~\ref{fig:frame number and BEV size}, we explore the impact of frame number and BEV resolution on performance. We can observe that the optimal BEV size $(W^b, H^b)$ is $512 \times 512$. 
Meanwhile, too small BEV resolution would cause the network to be unable to capture the motion of small objects, while excessively large resolution leads to sensitivity to slight disturbances. Besides, a larger BEV image will contain more numerous empty grids, which may dilute useful information.

Furthermore, as shown in Fig.~\ref{fig:frame number and BEV size}(b), compared to previous approaches~\cite{insmos,4DMOS} that require a lot of frames to extract the spatial-temporal features, our method only relies on 3 frames to achieve the best result. 
We think this is due to the effective reuse of historical feature and predictions in temporal fusion and voting, which bring rich prior knowledge to the network. 
Meanwhile, feeding too many frames into the network would cause information redundancy and result in degradation.

\section{Conclusion} 
In this paper, we analyze the limitations of existing MOS methods and propose a novel streaming structure, which uses memory bank as a bridge to transfer prior information among inferences while capturing the appearance and motion feature of objects from multiple views. To correct false predictions, we propose a voting mechanism to integrate historical results at the voxel and instance levels. Experimental results indicate that our method performs competitively in diverse aspects. 

\bibliographystyle{ieeetr}
\bibliography{Manuscript}

\end{document}